\documentclass[lettersize,journal]{IEEEtran}
\usepackage{amsmath,amsfonts}
\usepackage{algorithmic}
\usepackage{array}
\usepackage[caption=false,font=normalsize,labelfont=sf,textfont=sf]{subfig}
\usepackage{cite} 
\usepackage{textcomp}
\usepackage{stfloats}
\usepackage{url}
\usepackage{verbatim}
\usepackage{graphicx}
\usepackage{float}
\usepackage{booktabs}
\usepackage[hidelinks]{hyperref}
\usepackage{subfig}
\def\BibTeX{{\rm B\kern-.05em{\sc i\kern-.025em b}\kern-.08em
    T\kern-.1667em\lower.7ex\hbox{E}\kern-.125emX}}
\usepackage{balance}
\begin{document}
\title{FollowNet: A Comprehensive Benchmark for Car-Following Behavior Modeling}

\author{Xianda Chen, Meixin Zhu,~\IEEEmembership{Member,~IEEE,} Kehua Chen, Pengqin Wang, Hongliang Lu, Hui Zhong, Xu Han, Yinhai Wang,~\IEEEmembership{Fellow,~IEEE} 
\thanks{Manuscript created May 2023; This study is sponsored by Guangzhou Basic and Applied Basic Research Project (2023A03J0106) and Guangzhou Municipal Science and Technology Project (2023A03J0011). (Corresponding author: Meixin Zhu.)}
\thanks{Xianda Chen is with the Systems Hub, Intelligent Transportation Thrust, The Hong Kong University of Science and Technology (Guangzhou), Guangzhou, China}
\thanks{Meixin Zhu is with the Systems Hub, Intelligent Transportation Thrust,
Hong Kong University of Science and Technology (Guangzhou), Guangzhou, China; also with the Department of Civil and Environmental Engineering, The Hong Kong University of Science and Technology, Hong Kong; and Guangdong Provincial Key Lab of Integrated Communication, Sensing and Computation for Ubiquitous Internet of Things (e-mail: meixin@ust.hk).
}
\thanks{Kehua Chen is with the Division of Emerging Interdisciplinary Areas, The Hong Kong University of Science and Technology, Hong Kong, China}
\thanks{Pengqin Wang, Hongliang Lu, Hui Zhong and Xu Han are with The Hong Kong University of Science and Technology (Guangzhou), Guangzhou, China}

\thanks{Yinhai Wang is with Department
of Civil and Environmental Engineering, University of Washington, Seattle, 98195, Washington, United States, e-mail: yinhai@uw.edu}}


\maketitle
\begin{abstract}
Car-following is a control process in which a following vehicle (FV) adjusts its acceleration to keep a safe distance from the lead vehicle (LV). Recently, there has been a booming of data-driven models that enable more accurate modeling of car-following through real-world driving datasets. Although there are several public datasets available, their formats are not always consistent, making it challenging to determine the state-of-the-art models and how well a new model performs compared to existing ones. In contrast, research fields such as image recognition and object detection have benchmark datasets like ImageNet, Microsoft COCO, and KITTI. To address this gap and promote the development of microscopic traffic flow modeling, we establish a public benchmark dataset for car-following behavior modeling. The benchmark consists of more than 80K car-following events extracted from five public driving datasets using the same criteria. These events cover diverse situations including different road types, various weather conditions, and mixed traffic flows with autonomous vehicles. Moreover, to give an overview of current progress in car-following modeling, we implemented and tested representative baseline models with the benchmark. Results show that the deep deterministic policy gradient (DDPG) based model performs competitively with a lower MSE for spacing compared to traditional intelligent driver model (IDM) and Gazis-Herman-Rothery (GHR) models, and a smaller collision rate compared to fully connected neural network (NN) and long short-term memory (LSTM) models in most datasets. The established benchmark will provide researchers with consistent data formats and metrics for cross-comparing different car-following models, promoting the development of more accurate models. We open-source our dataset and implementation code in \url {https://github.com/HKUST-DRIVE-AI-LAB/FollowNet}. 

\end{abstract}


\section{Introduction}
\IEEEPARstart{C}{ar-following} is the most fundamental and frequent driving behavior. It involves actions taken by a driver when following another vehicle ahead. Proper car-following behavior can lower crash risks and improve traffic flow stability \cite{wang2016drivers, li2018situation, groelke2021predictive, wang2022effect}. The corresponding car-following model is a mathematical or computational representation of the behavior exhibited by drivers when following other vehicles on the road. It describes and predicts the dynamics of following vehicles (FV) and lead vehicles (LV) movements in traffic flow and is a cornerstone for microscopic traffic simulation \cite{gipps1981behavioural, panwai2005comparative}.

Over the past decade, there has been a boom of data-driven car-following models, primarily due to the availability of real-world driving data and advancements in machine learning. Representative data-driven car-following models include neural network based \cite{zheng2013car, chong2013rule}, recurrent neural network (RNN) based \cite{zhou2017recurrent, huang2018car}, and reinforcement learning (RL) based \cite{zhu2018human, zhu2020safe, tang2021atac, li2023modified}. However, existing research has the following limitations:
\begin{enumerate}
\item \textbf{Lack of standardized data formats and evaluation criterion.} Although there are multiple public driving datasets available (e.g., NGSIM, HighD), there are no standardized car-following datasets and evaluation metrics, making it difficult to compare newly proposed car-following models with existing ones in terms of performance.
\item \textbf{Inadequate representation of car-following behavior in mixed traffic flows.} 
We will likely face a transitional phase of mixed traffic where autonomous and human-driven vehicles share the road. However, previous studies have mainly focused on modeling car-following behavior using limited datasets that do not include autonomous vehicles.
\end{enumerate}

\begin{figure*}[htbp]
\centering
\includegraphics[width=1\linewidth]{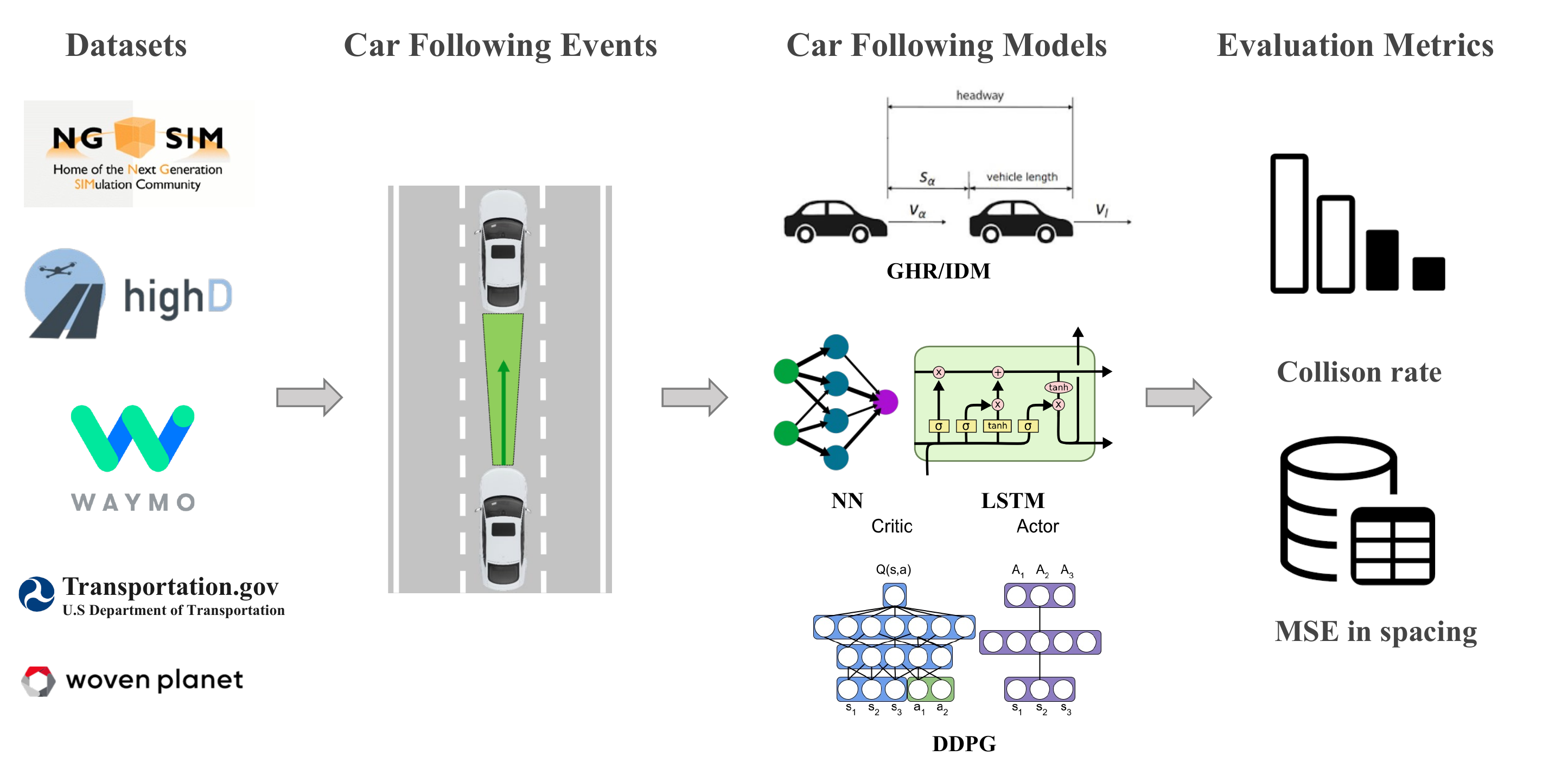}
\caption{A Roadmap of Car-following Benchmark: FollowNet}
\label{fig:overview}
\end{figure*}

To further advance the field of microscopic traffic simulation modeling, it is imperative to establish a public car-following benchmark that can address the aforementioned issues and serve as a standard dataset for research in car-following. Such a benchmark would be beneficial for advancing the field of microscopic traffic flow modeling, similar to how standard datasets such as ImageNet \cite{krizhevsky2017imagenet}, Microsoft COCO \cite{lin2014microsoft}, and KITTI \cite{Geiger2013IJRR} have contributed to their respective fields. To achieve this, we have created the benchmark by extracting data from five public datasets based on the same criteria. Our benchmark includes five baseline car-following models, consisting of both traditional and data-driven models (Figure. \ref{fig:overview}). This paper presents a summary of the recent developments in this field, as well as evaluates the performance of mainstream models using our benchmark.

The contributions of this study include:
\begin{itemize}
    \item Summarized existing car-following models systematically.
    \item Established the first benchmark of the car-following behavior to save the effort of data extraction and facilitate the development of car-following models. 
    \item Offered diverse scenarios for testing car-following models, including mixed traffic and different road types, while evaluating these models using standardized metrics.
    \item Provided references for the community on car-following extraction, calibration, and baseline model implementations, through data and code open-sourcing.
\end{itemize}

The rest of the paper is organized as follows. Section II reviews car-following models. Section III analyzes the characteristics of car-following events in different datasets. Section IV introduces five baseline models. Section V evaluates the model performance with consistent metrics. Section VI discusses future directions of car-following research. Section VII concludes the paper.

\section{Related Work}
\subsection{Traditional Car-Following Models}
Since the Pipes model was proposed in 1953 \cite{pipes1953operational}, researchers have studied car-following models for 70 years. In general, the existing traditional car-following models are divided into five main categories: stimulus-response model, safe distance model, psycho-physiological model, optimal velocity model, and desired goal model \cite{brackstone1999car, saifuzzaman2014incorporating, olstam2004comparison}.

The stimulus-response model uses the relative distance or relative speed to determine the acceleration of the FV. The Gazis-Herman-Rothery (GHR) model \cite{chandler1958traffic}, the first stimulus-response model, assumes that a vehicle's acceleration positively relates to the speed difference and that the FV's response is expressed regarding its acceleration or deceleration behavior. The basic equation of this model can be expressed as:
\begin{equation}
a_n(t)=c v_n^m(t) \frac{\Delta v(t-T)}{\Delta x^l(t-T)}
\end{equation}
where the $n$-th vehicle's acceleration at time $t$ is represented by $a_n(t)$, $\Delta v$, and $\Delta x$ are the speed difference and distance between the $(n-1)$ th vehicle and the $n$ th vehicle, respectively. $m$, $c$, and $l$ are constants that need to be calculated, and $T$ is the driver's reaction time. The GM car-following model \cite{gazis1961nonlinear} is another well-known stimulus-response model which used the difference between the LV and FV along with the velocity and space headway to determine the acceleration.

\cite{kometani1959dynamic} proposed the first safety distance model. The underlying principle of the model differs from the stimulus-response model due to the LV's unpredictable motion. The model is also known as the conflict avoidance model since the FV keeps a minimum safe distance even though the LV implements sudden breaks. Its basic formula is: 
\begin{equation}
\Delta x(t-T)=\alpha v_{n-1}^2(t-T)+\beta_l v_n^2(t)+\beta v_n(t)+b_0
\end{equation}
where $\Delta x$ denotes the distance between the $(n-1)$th vehicle and the $n$-th vehicle, $v_n(t)$ is the speed of the $n$-th vehicle at time $t$, $T$ is the driver's reaction time, $\alpha$, $\beta_l$, $\beta$, $b_0$ are parameters to be calibrated. Similarly, Gipps model \cite{gipps1981behavioural}  utilizes the concept of the safe distance to facilitate car-following behavior. Specifically, the driver of the FV selects a certain speed and maintains a corresponding distance from the LV to prevent a collision.

The psycho-physiological model suggests that drivers adopt strategies based on the relative motion between the LV and FV, including changes in speed and distance differences, and only react when the threshold value is exceeded. This model was first proposed in \cite{michaels1963perceptual}, and is used in VISSIM\ensuremath{^{{\textsuperscript{\textregistered}}}}, a microscopic traffic simulation tool, through the Wiedemann model \cite{wiedemann1974simulation}.

In \cite{bando1995dynamical}, the optimal velocity model (OVM) was introduced for the first time, which can account for various traffic flow phenomena, such as free flow, congested traffic, the relationship between density and traffic flow, and stop-and-go traffic waves. However, the OVM model may produce unrealistic acceleration and deceleration processes. To address this limitation, the generalized force (GF) model \cite{helbing1998generalized} was introduced, which added the effect of negative speed difference to the optimal velocity model. However, both models neglect situations where the speed of the LV is much slower than that of the FV. The full velocity difference (FVD) model \cite{jiang2001full} considers the safety distance and offers a more precise acceleration function, which is given by:
\begin{equation}
    \begin{array}{l}a_{n}(t)=\alpha\left[V_{n}^{*}\left(\Delta X_{n}(t)\right)-V_{n}(t)\right]+\lambda\left(\Delta V_{n}(t)\right) \\ \lambda=\left\{\begin{array}{ll}\lambda_{0}: & \Delta X_{n}(t) \leq s_{c} \\ 0: & \Delta X_{n}(t)>s_{c}\end{array}\right.\end{array}
\end{equation}
The first term in the acceleration function is proportional to the difference between the optimal velocity $V_{n}^{*}\left(\Delta X_{n}(t)\right)$ and the actual velocity $V_{n}(t)$, and the second term considers the velocity difference $\Delta V_{n}(t)$ as a linear stimulus. The sensitivity coefficients are denoted by $\alpha$ and $\lambda$, and $s_{c}$ is a threshold value that distinguishes between following and free driving. 


The desired goal model assumes that the driver has certain desired goals, such as desired following speed, desired headway, etc. The Intelligent Driver Model (IDM) \cite{treiber2000congested} is the most frequently used driver-based desired goal model. It is also by far the most thorough and succinct theoretical car-following model that is free of accidents which assumes that each driver has a set of desired parameter values during a period of following behavior trying to maintain. The model expressions are:
\begin{equation}
a_n(t)=a_0\left[1-\left(\frac{v_n(t)}{\widetilde{v}_n}\right)^\lambda-\left(\frac{\widetilde{S}_n(t)}{s_n}\right)^2\right] 
\end{equation}
\begin{equation}
\widetilde{S}_n(t)=S_0+v_n(t) \widetilde{T}+\frac{v_n(t) \Delta v(t)}{2 \sqrt{a_0 b}} 
\end{equation}
\begin{equation}
\Delta v(t)=v_n(t)-v_{n-1}(t)
\end{equation}
where $a_n(t)$ and $v_n(t)$ represent the acceleration and the velocity of the FV at time $t$, respectively. And $S_n(t)$, $\Delta v(t)$ are the spacing and relative speed between the FV and the LV. The desired maximum acceleration, maximum deceleration, desired velocity, and desired time headway are represented by $a_0$, $b$, $\widetilde v$, and $\widetilde{T}$, respectively. $S_0$ is the minimum safe headway and $\lambda$ is a constant to be calibrated. Subsequently, the Intelligent Driver Model with Memory (IDMM) \cite{treiber2003memory} is introduced which designs the IDM to incorporate memory effects and adapt driving behavior to the surrounding traffic. 

Traditional car-following models have several disadvantages, which can be summarized into three main points:
\begin{enumerate}
    \item \textbf{Simplistic and unrealistic assumptions.} Many traditional car-following models oversimplify driver behavior and do not reflect real-world conditions. 
    \item \textbf{Limited adaptability and data availability.} Traditional car-following models may lack the ability to adapt to dynamic traffic flow, and their effectiveness may be limited by the amount of data available.
    \item \textbf{Limited applicability.} Traditional car-following models are often only applicable to a specific type of road and may not be able to capture the complexity of mixed-use road networks or account for emerging driving technologies beyond human-driver behavior.
\end{enumerate}
More advanced car-following models have been developed that are able to get around some of these limitations as a result of recent advancements in technology as well as data collection techniques. However, in the study of traffic engineering and flow analysis, traditional ones are still very important and frequently employed.

\begin{figure*}[htbp]
\centering
\includegraphics[width=0.8\linewidth]{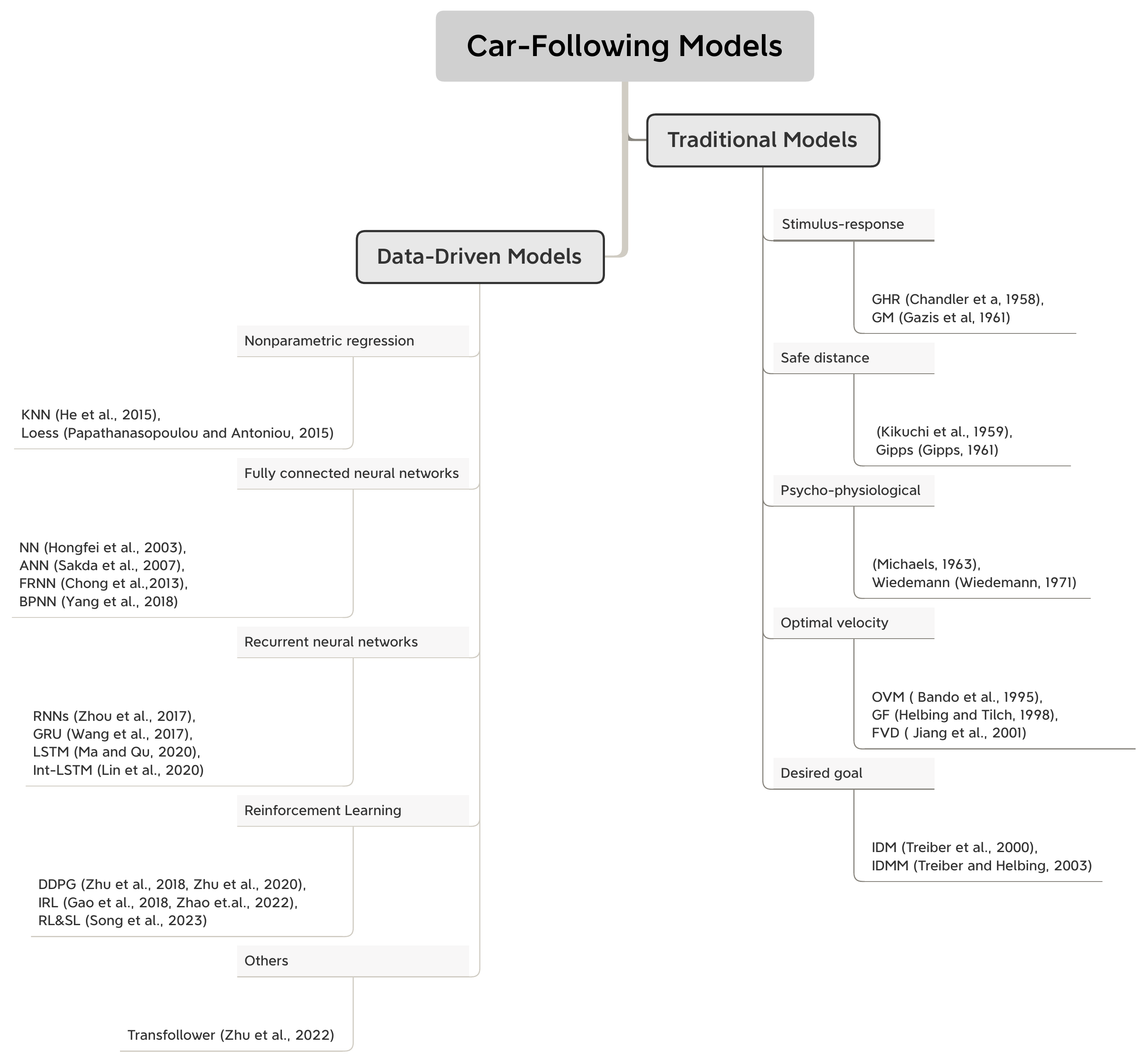}
\caption{The development of car-following models}
\label{fig:ALL}
\end{figure*}

\subsection{Data-Driven Car-Following Models}
Traditional models and data-driven models are the two main categories of car-following models, as shown in Figure. \ref{fig:ALL}. Data-driven models make use of artificial intelligence techniques such as nonparametric regression, fully connected neural networks, recurrent neural networks, reinforcement learning, and other methods to predict drivers' behavior. These models learn relationships between different factors and the driver's behavior from the collected data.

A straightforward k-nearest neighbor (KNN)-based nonparametric car-following model \cite{he2015simple} was presented in 2015 that forecasts the most likely driving behavior under the given circumstances. Similarly, Loess model \cite{papathanasopoulou2015towards} is also a nonparametric data-driven car-following model based on locally weighted regression.

Another common approach to predict the acceleration of an FV is to use a fully connected neural network (NN). A 4-layer neural network with two hidden layers \cite{hongfei2003develop} takes the gap distance, the relative velocity, desired velocity, and follower velocity as inputs to predict the FV's acceleration directly. In \cite{panwai2007neural}, the reactive agent-based car-following models were developed using artificial neural network (ANN) techniques, including backpropagation and fuzzy ARTMAP architectures. The results showed that the models' degrees of accuracy were best for the backpropagation and fuzzy ARTMAP architectures. Also, in \cite{chong2013rule}, one hidden layer neural network can accurately predict acceleration and come up with the fuzzy rule-based neural network (FRNN), which was further improved by giving the neural network an instantaneous reaction time (RT) delay \cite{khodayari2012modified}. In \cite{yang2018novel}, a Back-Propagation Neural Networks (BPNN) model combined with the Gipps model is proposed to avoid crashes.

Recurrent neural networks are another popular network structure that can model temporal problems such as vehicle following using historical information. Vanilla RNN \cite{zhou2017recurrent} was utilized to simulate drivers' car-following behavior and demonstrated effective traffic oscillation prediction. Two other popular RNN variants, Long Short-Term Memory (LSTM) and Gated Recurrent Unit (GRU) have also been used to model car-following behavior. \cite{ma2020sequence} utilized LSTM to address issues of gradient vanishing and exploding during long sequence training, resulting in improved performance of the car-following model. \cite{wang2017capturing} used GRU to simulate car-following behavior to improve training efficiency. To address the issue of error propagation, \cite{lin2020platoon} suggested an LSTM-based interconnected car-following model (Int-LSTM).

In a reinforcement learning approach, the agent learns an optimal control policy through trial and error in an unknown environment based on a reward function. \cite{zhu2018human} presented a framework for automatic car-following planning based on deep RL, which aimed to accurately reproduce human-like car-following behavior. However, human driving may not be the optimal driving operation \cite{chai2015fuzzy}, car-following behavior should be optimized in terms of safety, efficiency, and comfort rather than simply imitating human drivers. \cite{zhu2020safe} proposed a reward function that aimed to achieve two objectives: mimicking human drivers and optimizing driving performance. Driving characteristics and human driving data were used to design the reward function and the agents were trained to learn the decision mechanism by continuously utilizing history-driven information. Other studies \cite{gao2018car,zhao2022personalized, song2023personalized} assessed each driver's driving characteristics and car-following behaviors using inverse reinforcement learning (IRL) or the combination of RL and Supervised Learning (SL).

In addition, a long-sequence car-following trajectory prediction model based on the Transformer attention-based model was proposed in \cite{zhu2022transfollower}, which is a typical encoder-decoder architecture. The encoder uses multi-head self-attention to create a mixed representation of the past driving environment utilizing historical spacing and speed data as inputs. This means that the model can effectively capture complex temporal relationships between the data, allowing it to produce a more accurate representation of the driving context.

Despite advances in data-driven car-following models, they still lack interpretability and generalization ability. Developing more interpretable and generalizable mechanism models can address these limitations. Also, standardized testing datasets and evaluation criteria are needed to compare performance and determine the best model. While many public datasets are available for evaluation purposes, their data formats and standards may differ, and significant effort is required to familiarize the data structure of each dataset and extract car-following events. Therefore, creating a car-following benchmark among public datasets can simplify testing and foster microscopic traffic research development.

\section{Datasets and Car-Following Events }
\begin{figure*}[htbp]
    \centering
    \subfloat[HighD]{\includegraphics[width=0.31\textwidth]{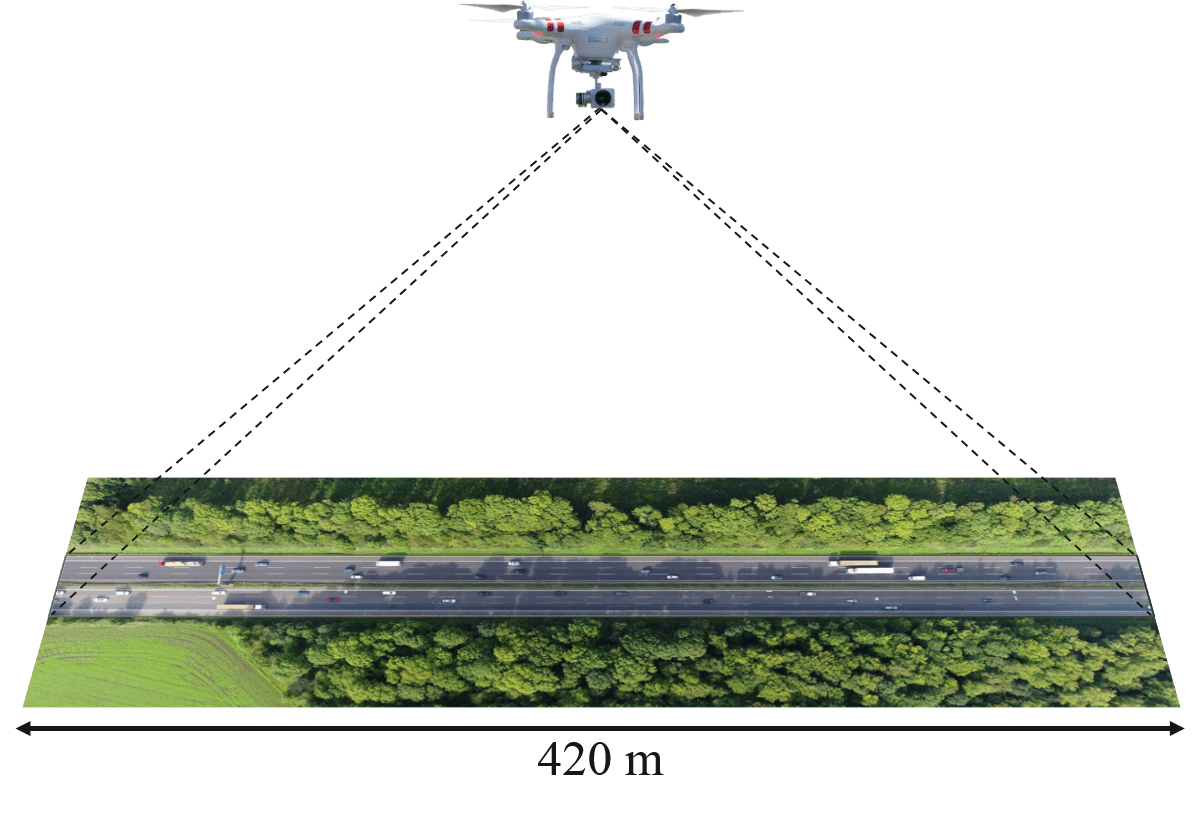}}\hspace{0.5pc}
    \subfloat[NGSIM]{\includegraphics[width = 0.26\textwidth]{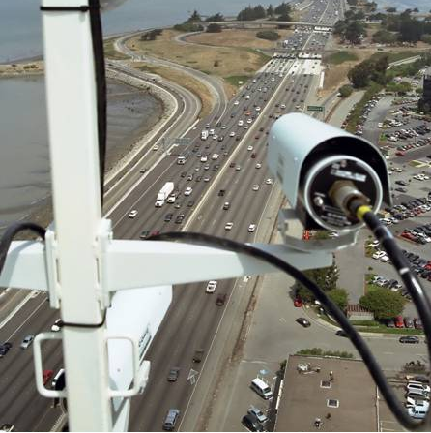}}\hspace{0.5pc}
    \subfloat[SPMD]{\includegraphics[width = 0.26\textwidth]{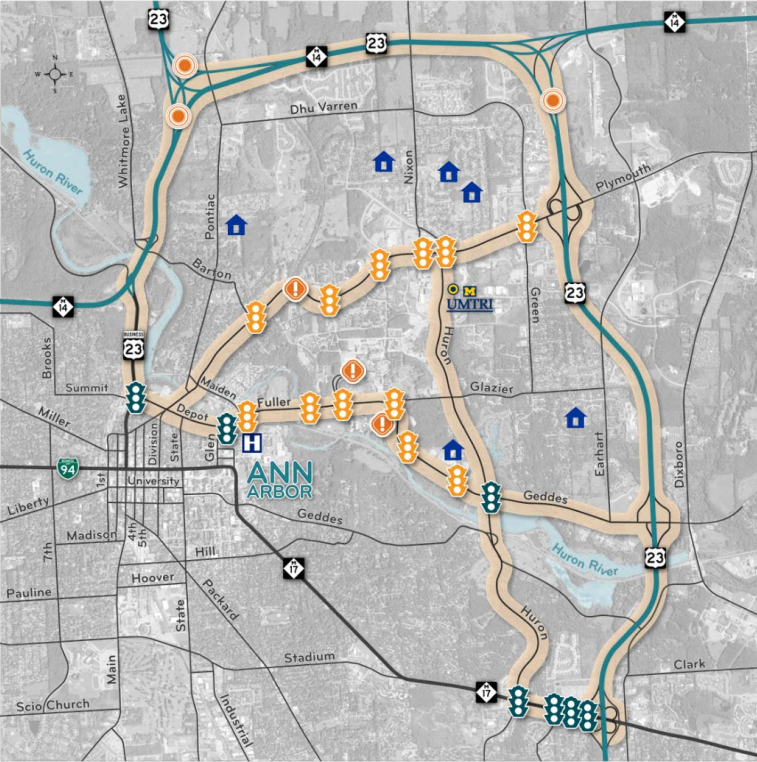}}\\
    \subfloat[Waymo]{\includegraphics[width = 0.4\textwidth]{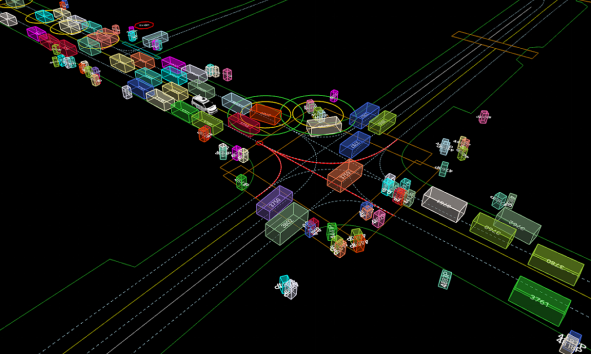}}\hspace{2.2pc}
    \subfloat[Lyft]{\includegraphics[width = 0.4\textwidth]{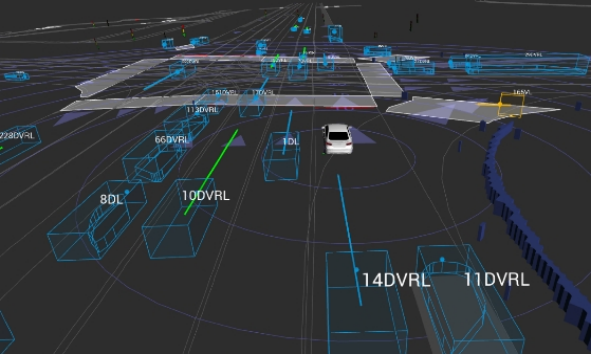}}
    \caption{Five Public Datasets.}
    \label{fig:dataset}
\end{figure*}

Several large-scale datasets have been created to aid in the development and evaluation of autonomous driving algorithms and technologies. As autonomous vehicles become more prevalent on the road, human trust in their capabilities can affect car-following behavior, among other variables. It is crucial to thoroughly investigate and understand these behaviors to ensure safe and efficient mixed traffic flow. This section extracts and summarizes the car-following events from five mainstream public datasets, as shown in Figure. \ref{fig:dataset}, where events in datasets Waymo and Lyft include AV (Autonomous Vehicles). The car-following events that lasted for more than 15 seconds have been recorded.
Additional filtering rules are applied depending on the characteristics of each dataset, as presented later in this paper. Table. \ref{tab:dataset} lists five commonly used datasets in the field of autonomous driving.
\begin{table*}[]
\caption{Five commonly used datasets in the field of autonomous driving.}
\label{tab:dataset}
\centering
\begin{tabular}{@{}llllll@{}}
\toprule
Dataset & Type                       & Scence                & Sensors      & AV involved & Data Hz \\ \midrule
HighD   & Vehicle                    & Expressway            & Camera       & No          & 25   \\  
NGSIM   & Vehicle                    & Expressway, Urban street & Camera       & No          & 10            \\
SPMD (DAS1, DAS2)    & Vehicle                    & Urban street           & Camera, GPS   & No          & 10                   \\
Waymo   & Vehicle, Bicycle & Urban street            & Camera, Lidar & Yes       & 10            \\
Lyft    & Vehicle, Bicycle & Urban street            & Camera, Lidar & Yes        & 10            \\
\bottomrule
\end{tabular}
\end{table*}
\subsection{HighD}
The HighD dataset \cite{highDdataset} released by the Institute of Automotive Engineering at RWTH Aachen University in Germany is a comprehensive driving dataset that provides high-precision information about vehicle positions and speeds. The dataset includes bird's-eye view videos of six different roads around Cologne, Germany, captured by a high-resolution 4K camera mounted on an aerial drone. The vehicle position and speed information are extracted using advanced computer vision techniques, which ensure that the positioning error is under 10 cm. To ensure that the recorded events are representative and comparable, car-following events with prolonged periods of low speed or stoppage are filtered out by excluding those with average speeds below 2 m/s or speeds below 0.2 m/s that last for more than 5 seconds.

\subsection{NGSIM}
NGSIM (Next Generation Simulation) is a traffic dataset created by the Federal Highway Administration (FHWA) to study the dynamics of traffic flow on freeways. Four different scenarios are publicly available: US101 in Los Angeles, California; southbound of Lankershim Boulevard; Peachtree Street in Atlanta, Georgia and eastbound of I-80 in Emeryville, California, where the trajectory data is collected by simultaneous photography from cameras set up on the high ground. Among them, the I-80 section is located before the evening peak between 4:00 p.m. and 4:15 p.m., and the traffic condition is relatively smooth. Car-following events of I-80 are recovered based on the rebuilt data \cite{369980:8164268} because there is some noise in the original data.

\subsection{SPMD}
In 2014, the U.S. Department of Transportation (USDOT) conducted the Safety Pilot Model Deployment (SPMD) to evaluate a dedicated short-range communication technology for V2V safety applications. The data set includes basic safety messages, vehicle trajectory and various driver-vehicle interaction data, and weather data at the time of data collection. Most of the data presented in this dataset was gathered from vehicles that have both vehicle-to-infrastructure (V2I) and vehicle-to-vehicle (V2V) communication devices installed, in addition to several roadside sensors. Although vehicle sensing devices were installed on the majority of the vehicle test vehicles, there was no feedback on driving from the information obtained, so the driver could still be considered to be driving under natural conditions. Using the data acquisition system (DAS), the position and speed information of vehicles were extracted from the two driving datasets (DAS1, DAS2). Since the spacing data can be obtained not only from sensors directly, but also theoretically inferred from the relative speed simulation of two consecutive frames, some noisy data can be filtered by comparing these two spacing data.

\subsection{Waymo}
In August 2019, Waymo, a self-driving car company operated by Alphabet, the parent company of Google, announced the release of the Waymo Open Dataset \cite{llc2019waymo}. It includes high-resolution sensor data from lidar and cameras, along with precise vehicle poses and annotations of objects in 3D. The dataset contains various driving scenarios, such as urban, suburban, and highway driving, and has a total of 1950 scenes, each lasting 20 seconds. \cite{hu2022processing} manually extracted the car-following events from this dataset perception part and made the extracted dataset publicly available. On their basis, a total of 1440 following events were extracted.

\subsection{Lyft}
The Lyft autonomous driving dataset \cite{kesten2019lyft} is a valuable resource for researchers seeking to develop autonomous driving systems. It is a Level 5 autonomous driving dataset, which includes a high-definition spatial semantic map, over 55,000 3D artificial annotation frames, and data collected from 7 cameras and 3 LiDARs. This dataset is particularly useful for understanding car-following behavior in mixed traffic flows as it includes more than 170,000 scenarios of human-driven vehicles following autonomous vehicles and autonomous vehicles following human-driven vehicles. Each scene is recorded for approximately 25 seconds. To ensure that the recorded data is representative and comparable, car-following events involving prolonged periods of low speed or stoppage are filtered out.

To address the issue of noise in the initial data when applying datasets, we used the Savitzky-Golay filter for smoothing. This filter is a finite impulse response (FIR) filter that employs polynomial fitting to estimate the underlying signal and mitigate the effect of noise in the time series data.

\subsection{Car-Following Event Extract}
The present study extracted car-following events from five datasets utilizing a methodology similar to prior investigations \cite{zhu2018human, wang2017capturing, zhao2017trafficnet}. The selection criteria, outlined below, are applied:

\begin{itemize}
    \item In order to guarantee that the FV follows the same vehicle throughout the car-following event, the LV number should be continuous and unchanged;
    \item The duration of the event is greater than or equal to 15 seconds, thereby providing sufficient data for analysis;
    \item The lateral distance between the FV and LV is equal to or less than 2 meters, ensuring that they are driving in the same lane.
\end{itemize}

A total of 1930, 16658, 24247, 1440, 24093, and 12540 car-following events were extracted from the NGSIM I-80, SPMD (DAS1), SPMD (DAS2), Waymo, Lyft and HighD datasets, respectively. Regarding the data shape format for each dataset, it consists of the number of car-following events and 4 dimension data, which are spacing, following vehicle speed, relative speed, and lead vehicle speed. Each dimension has a specific time length. Furthermore, every dataset was divided into three parts: training, validation, and test, with respective amounts of 70\%, 15\%, and 15\%.

\subsection{Descriptive Statistics and Distributions of Behavioral Measures}
\begin{figure*}[htbp]
    \centering
    \subfloat[Space gap]{\includegraphics[width=0.32\linewidth]{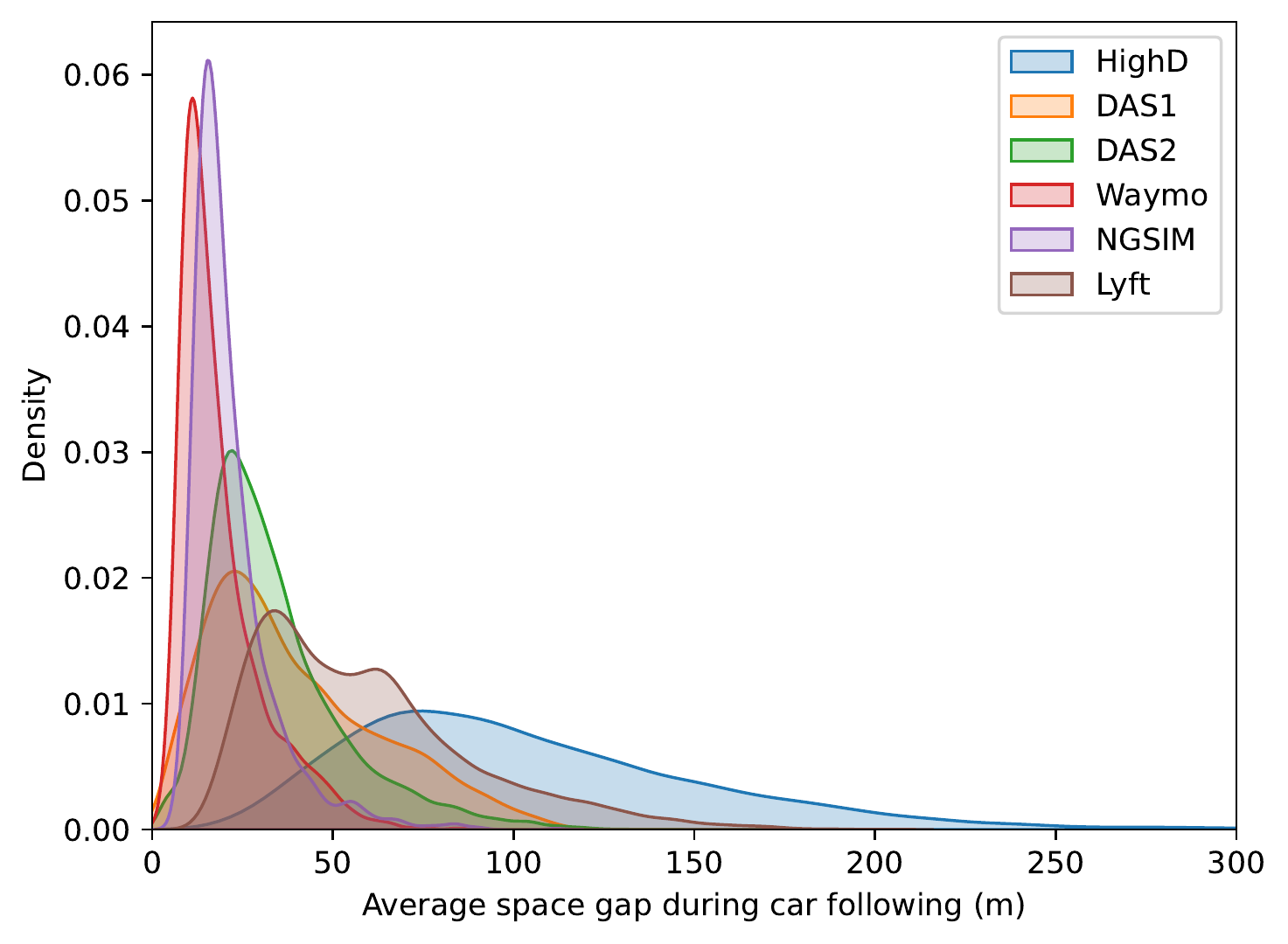}}
    \subfloat[Speed]{  \includegraphics[width=0.315\linewidth]{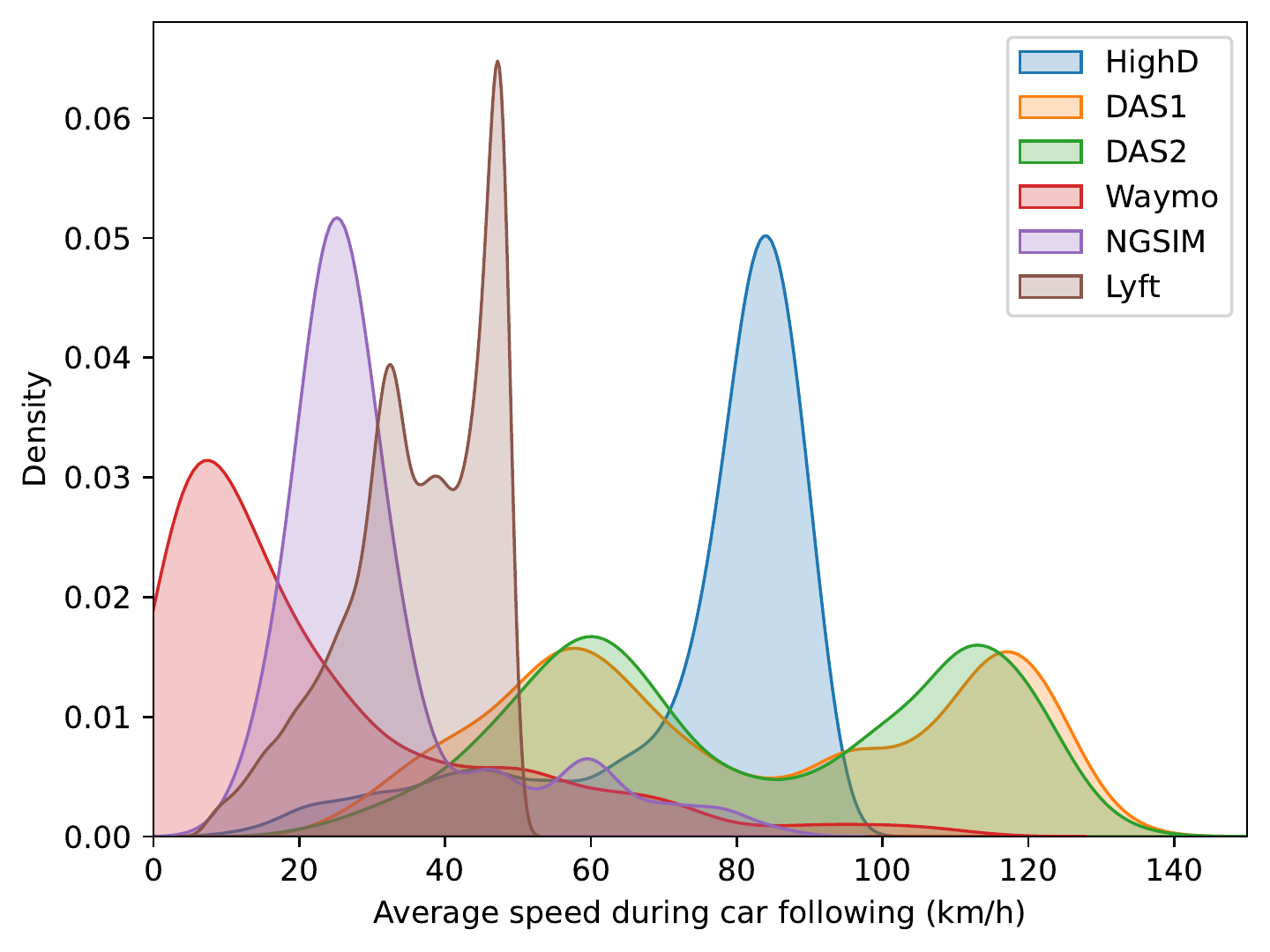}}
    \subfloat[Time gap]{ \includegraphics[width=0.32\linewidth]{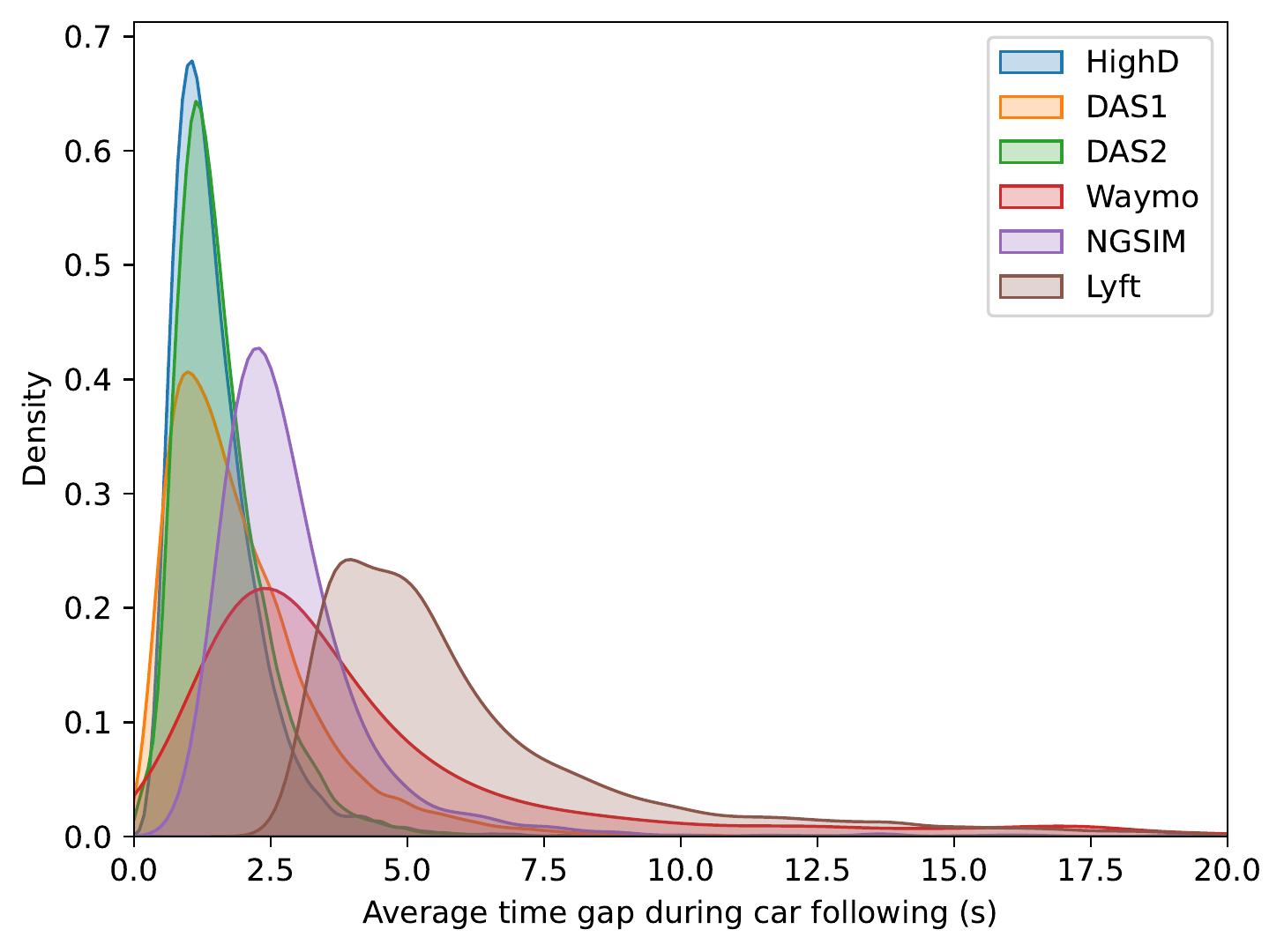}}\\
    \subfloat[Absolute relative speed]{\includegraphics[width=0.32\linewidth]{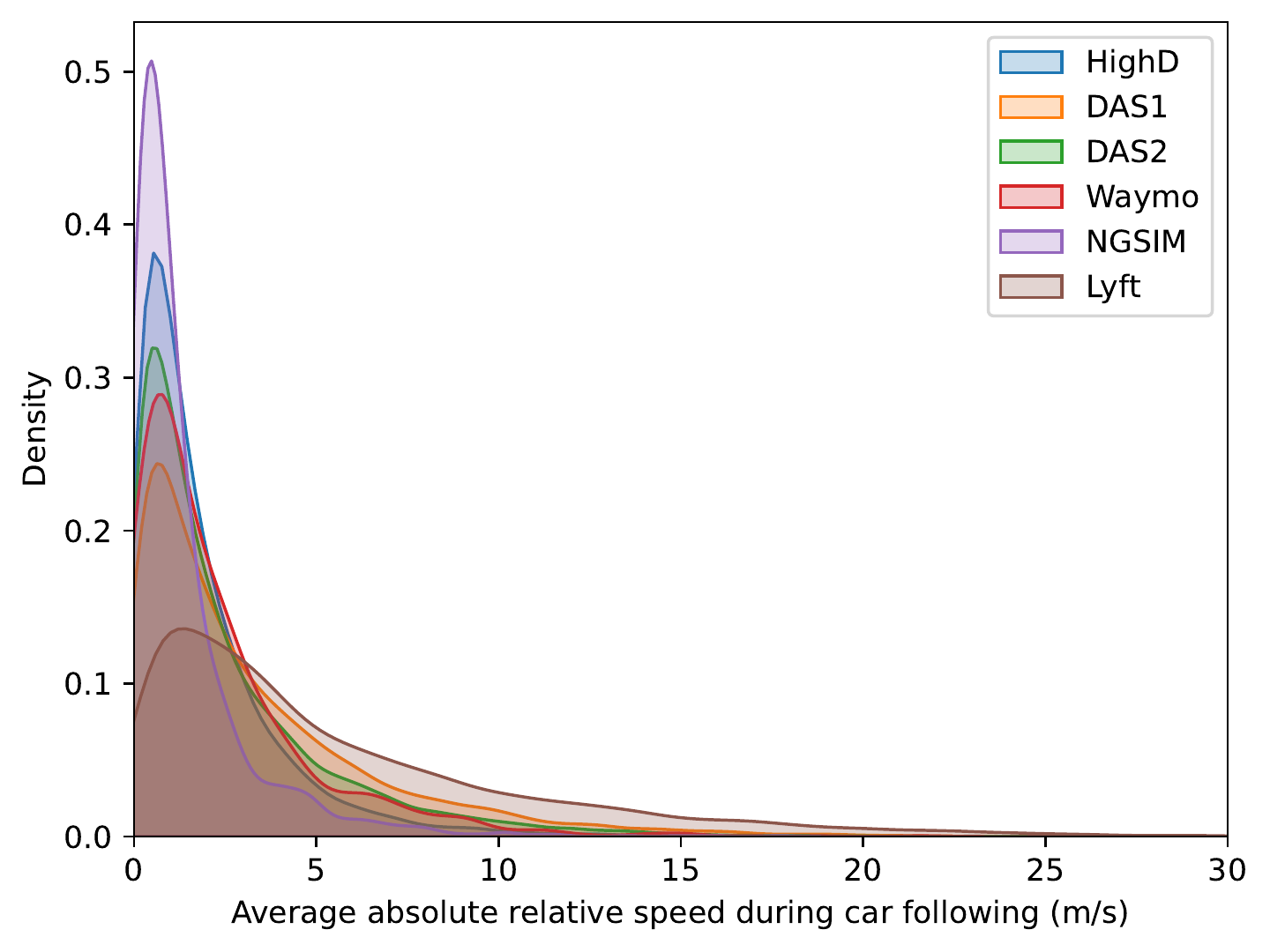}}\hspace{0.1pc}
    \subfloat[Absolute acceleration]{\includegraphics[width=0.32\linewidth]{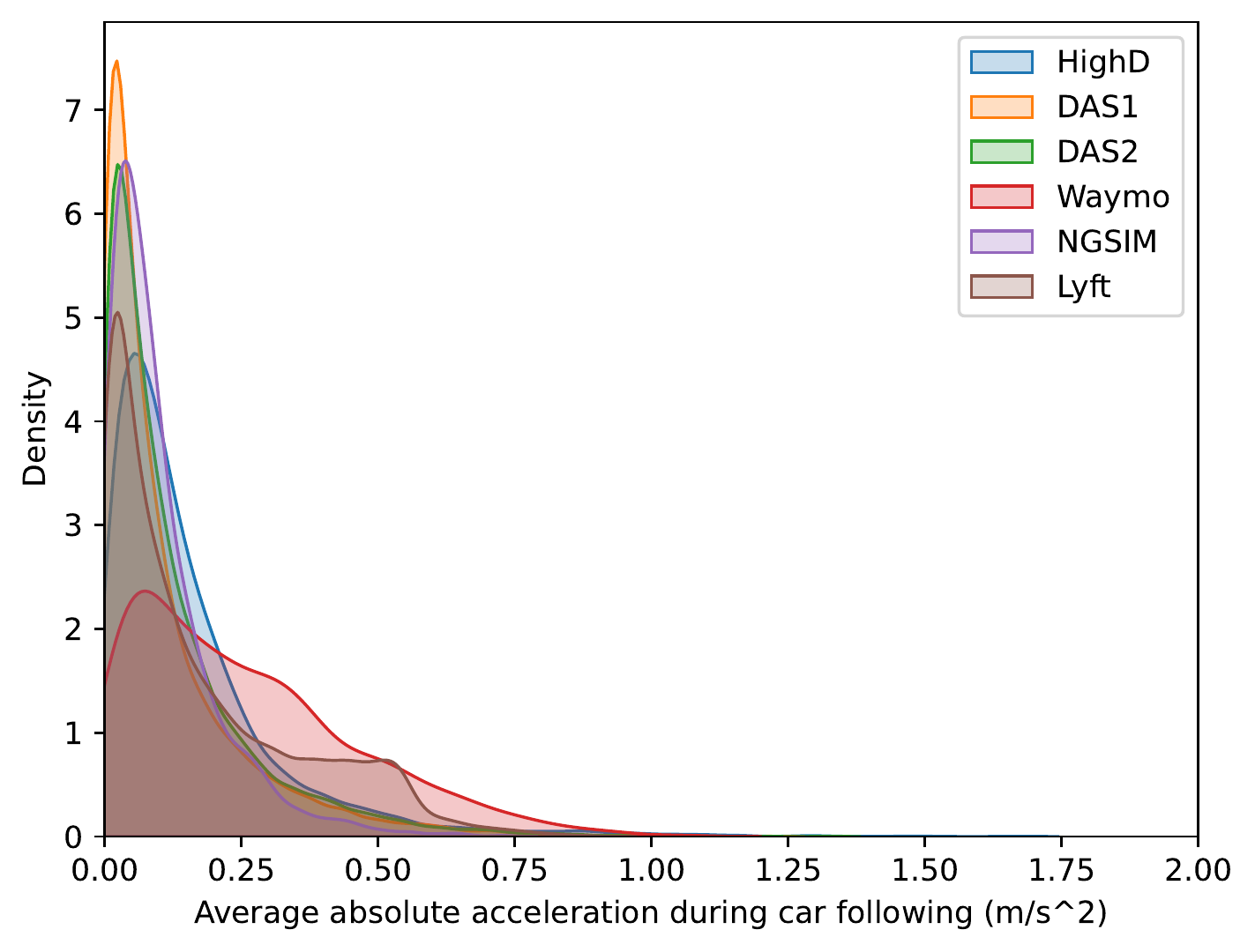}}
    \subfloat[Event duration]{\includegraphics[width=0.315\linewidth]{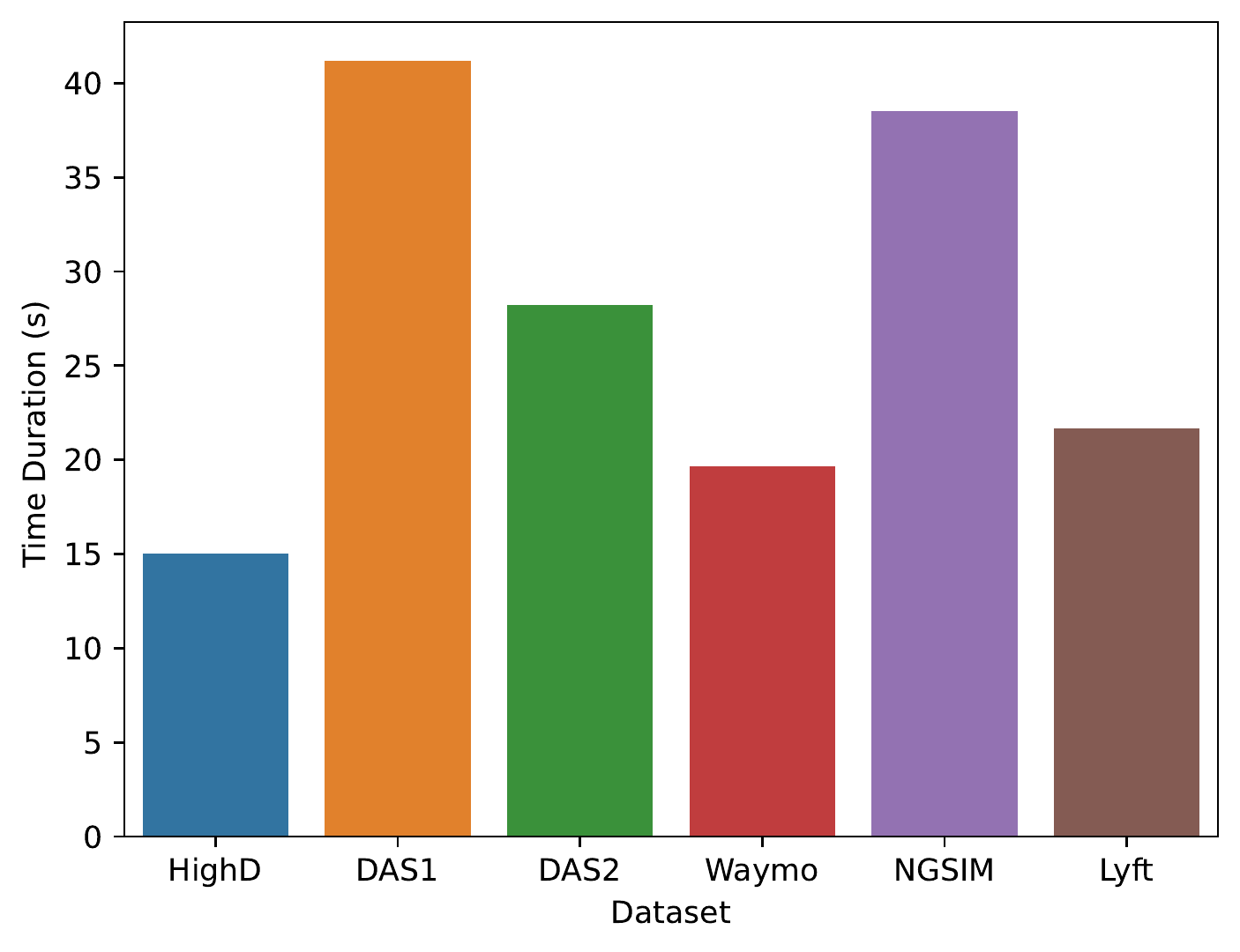}}    
    \caption{Distributions of car-following behavioral measurements.}
        \label{fig:distribution}
\end{figure*}

This section aims to present the descriptive statistics and distributions of six common car-following behavioral measures in the above datasets, as shown in Figure. \ref{fig:distribution}. We analyzed five public datasets of autonomous driving and studied six dimensions of car-following behavior. Our analysis shows that each dataset has unique characteristics related to its collection location and driving scenarios. These findings are crucial for developing safe and efficient autonomous driving systems capable of operating in various driving conditions. Below are the key findings from the comparison of car-following behavior across different datasets:
\begin{itemize}
    \item \textbf{Space gap}: HighD dataset has the largest average spacing gap, which is in line with the data collection occurring on highways where vehicles tend to maintain a larger following distance. On the other hand, Waymo dataset has the smallest average spacing gap, showing that the autonomous vehicles in this dataset are maintaining a smaller following distance, likely due to driving in more congested urban environments. DAS1 and DAS2 datasets have similar spacing gap distributions with an average of around 25 meters, indicating that vehicles in these datasets are also maintaining a relatively large space gap. The spacing gap is important in evaluating car-following behavior and can help develop guidelines for setting safe following distances for autonomous vehicles.  
    \item \textbf{Following speed}: the DAS1 and DAS2 datasets have relatively high following speeds, with data distributions that are centered around 60km/h and 110km/h. The Waymo dataset, on the other hand, has the lowest following speeds, as the car-following events primarily occur in low-speed driving scenarios in urban streets. However, HighD exhibits high following speeds, with data distributions that are centered around 80km/h, which is related to the dataset's highway driving scenario. The following speed is influenced by various factors such as traffic conditions, road type, and driver behavior. By analyzing the variations in following speeds across different datasets, researchers can gain valuable insights into how vehicles operate under different road and traffic conditions, which can be used to develop more effective algorithms for autonomous driving systems.
    \item \textbf{Time gap}: Lyft dataset has the largest time gap compared to other datasets, which suggests that the autonomous vehicles equipped with ADAS are maintaining a larger following distance from other vehicles compared to other datasets where human drivers may be more likely to maintain a shorter following distance. On the other hand, the SPMD (DAS1), SPMD (DAS2), and HighD datasets have smaller time gaps that are centered around one second, which may be attributed to their collection locations. This could imply that the driving behavior of the motorists in these locations encourages shorter following distances.
    \item \textbf{Absolute relative speed}: Most datasets exhibit similar data distributions for absolute relative speed suggesting that there may be some common driving behaviors across different locations and traffic conditions. However, the Lyft dataset features a different distribution, with most car-following events exhibiting absolute relative speeds in the range of 1 m/s. 
    \item \textbf{Absolute acceleration}: both Waymo and Lyft datasets exhibit a higher absolute acceleration distribution compared to other datasets, which could be attributed to the ADAS-equipped vehicles in these datasets having different acceleration patterns.
    \item \textbf{Car-following duration}: The car-following duration in the HighD is set to 15 seconds, while the NGSIM and DAS1 have an average car-following duration exceeding 35 seconds. This provides valuable data support for studying long-term car-following behavior.
    
\end{itemize}

\section{Baselines}

\begin{figure*}[htbp]
    \centering
    \subfloat[MSE for spacing]{\includegraphics[width=0.45\linewidth]{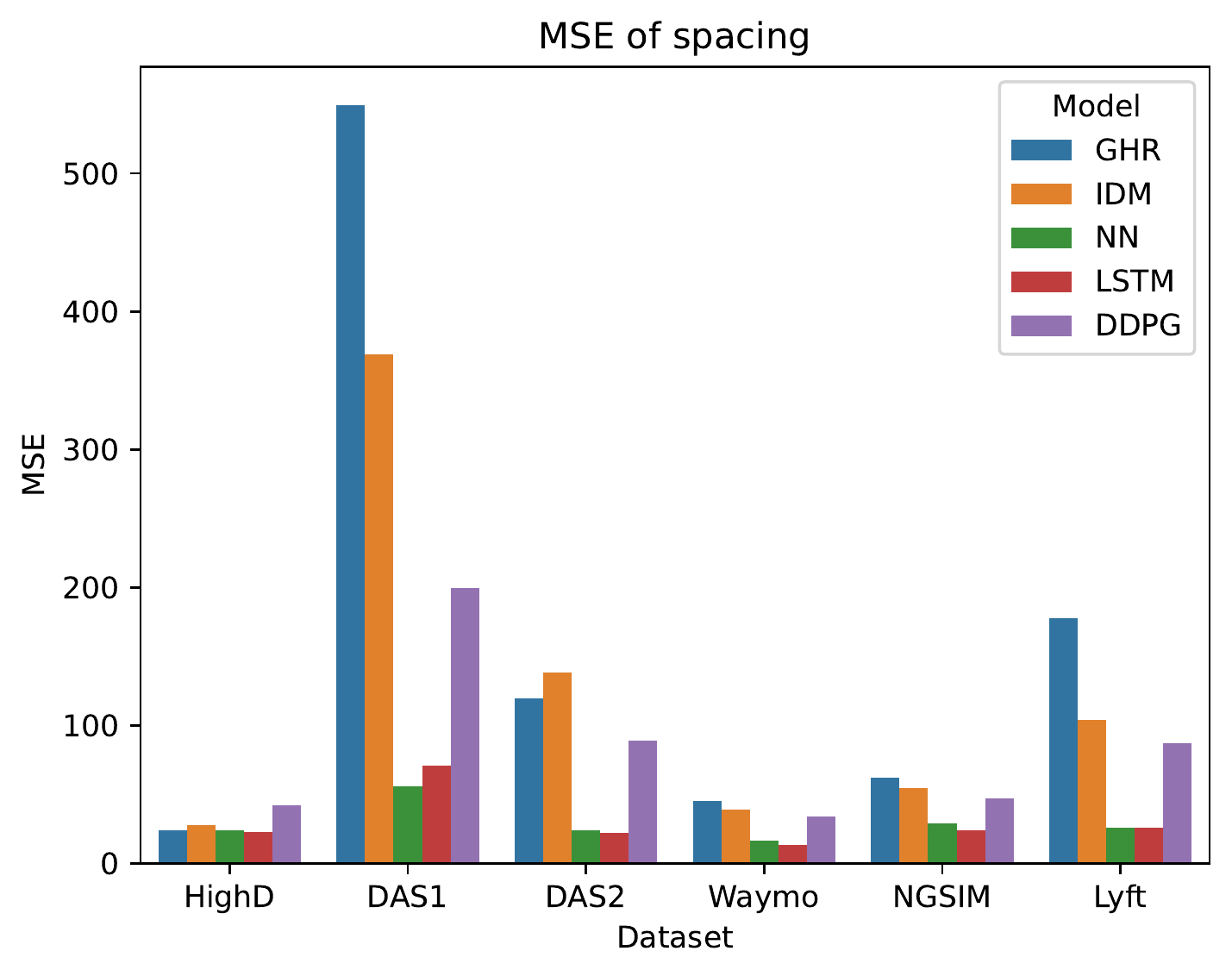}}
    \subfloat[Collison rate]{\includegraphics[width=0.45\linewidth]{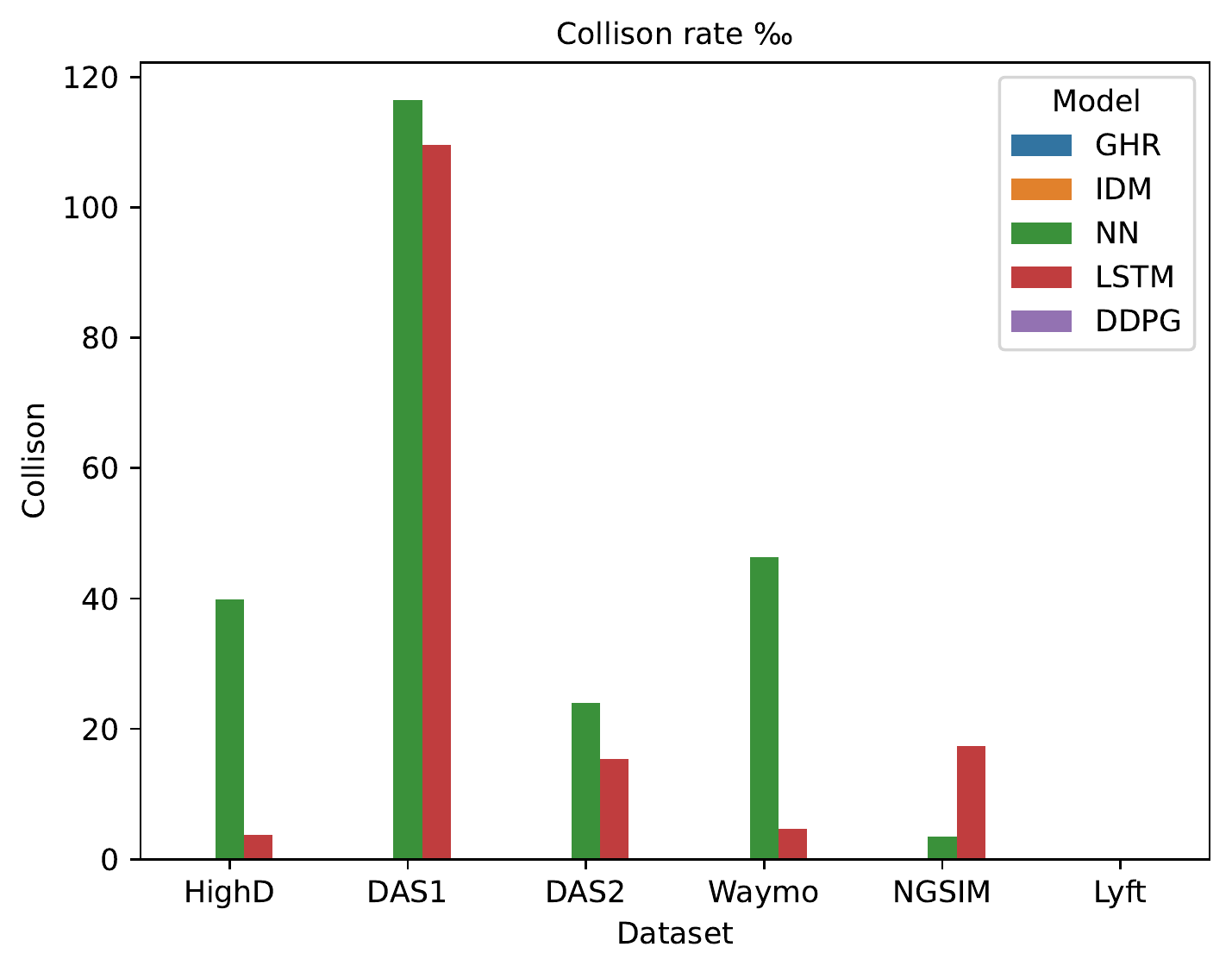}}   
         \caption{Model Benchmark Performance.}
             \label{fig:result}
\end{figure*}

\begin{table*}[]
\centering
\caption{Model Benchmark Performance: MSE of spacing and Collison rate }
\label{tab:results}
\begin{tabular}{@{}l|llllll|llllll@{}}
\toprule
      & \multicolumn{6}{l|}{MSE of spacing}              & \multicolumn{6}{l}{Collison rate \textperthousand ~ (Number of Collision)}                                      \\ \midrule
Model & HighD & DAS1   & DAS2   & Waymo & NGSIM & Lyft   & HighD      & DAS1         & DAS2       & Waymo      & NGSIM     & Lyft \\ \midrule
GHR   & 23.76 & 549.72 & 119.92 & 45.52 & 62.18 & 178.01 & 0          & 0            & 0          & 0          & 0         & 0    \\
IDM   & 27.50 & 368.95 & 138.69 & 39.17 & 54.97 & 104.32 & 0          & 0            & 0          & 0          & 0         & 0    \\
NN    & 24.06 & 56.01  & 24.14  & 16.79 & 29.00 & 26.06  & 39.87 (75) & 116.44 (291) & 23.92 (87) & 46.29 (10) & 3.46 (1)  & 0    \\
LSTM  & 22.90 & 70.69  & 22.49  & 13.75 & 23.86 & 25.71  & 3.72 (7)   & 109.64 (274) & 15.39 (56) & 4.62 (1)   & 17.30 (5) & 0    \\
DDPG  & 42.29 & 199.41 & 89.34  & 33.91 & 46.93 & 86.98  & \textbf{0} & \textbf{0}   & \textbf{0} & \textbf{0} & \textbf{0}&\textbf{0}\\ 
\bottomrule
\end{tabular}
\end{table*}

\begin{figure}[htbp]
	\centering
  \includegraphics[width=1\linewidth]{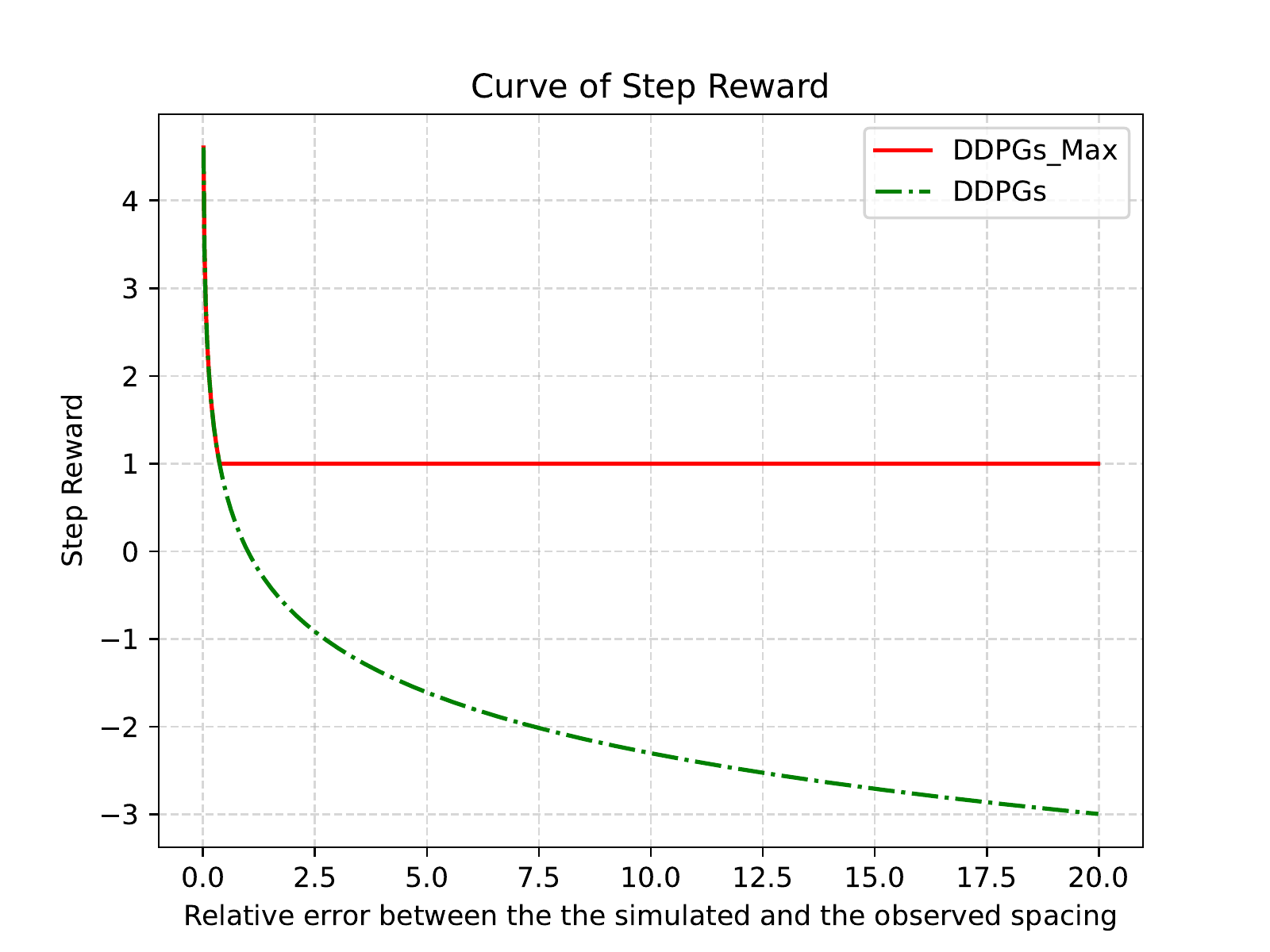}
                 \caption{Reward function of DDPGs\_Max and DDPGs.}
                         \label{fig:max}
\end{figure}

The five driving datasets mentioned above underwent training and testing procedures by utilizing two traditional car-following models as well as three data-driven models. The time interval for car-following events was set at 15 seconds. The five baseline models are listed as follows:
\begin{itemize}
    \item \textbf{IDM}: IDM demonstrates the best predictive performance when compared with other traditional models according to \cite{zhu2018modeling}. We proceeded to train an IDM model with the objective of minimizing the Mean Squared Error (MSE) of spacing by employing the genetic algorithm (GA) to determine the most effective IDM parameter set.

    \item \textbf{GHR}: The underlying premise of the GHR model is that the FV determines its acceleration by contrasting its speed to that of the LV. Similar to the IDM, GA was utilized to discover the optimal parameter values. (see \cite{zhu2018modeling} for calibration details). 

    \item \textbf{Fully connected neural network (NN)}: To forecast the future acceleration of the FV, a neural network model consisting of three feedforward layers with hyperbolic tangent (Tanh) activation functions is employed. The Adam optimizer with a learning rate of 0.001 is utilized for optimizing all models. Following this, the MSE of spacing loss function is utilized for optimizing the network.
    
    \item \textbf{Long Short-Term Memory (LSTM)}: To optimize the LSTM model's performance, we adjusted hyperparameters such as hidden size and LSTM layers, and a dropout probability of 0.1 is employed to enhance the robustness of the model.
    
    \item \textbf{Deep Deterministic Policy Gradient (DDPG)}: In \cite{zhu2018human}, DDPGs has been proved to be effective to minimize the discrepancy between the simulated and actual actions in order to reproduce human-like driving. However, the proposed reward function can not generalize to all datasets. Therefore, a modified reward function ${r_t}$ (DDPGs\_Max) with max operation and collision penalty is proposed in this paper, which can achieve less collision and higher accuracy. It be expressed as below:
    
\begin{equation}  \begin{split} {r_t} = -\max(\log \left( {\left| {\frac{{{S_{n - 1,n}}(t) - S_{n - 1,n}^{obs}(t)}}{{S_{n - 1,n}^{obs}(t)}}} \right|} \right),H) \\  +CollisionCheck*Penalty\end{split}\end{equation}

where ${S_{n - 1,n}}(t)$ and $S_{n - 1,n}^{obs}(t)$ are the simulated and observed spacing at time step $t$, respectively. And $H$ is the step reward, which is an adjustable parameter, set to 1 here. The collision check with penalty is set to let the agent learn in the direction of collision reduction in the RL environment. The condition for the end of training is that a collision occurs or the exploration of the entire car-following event is completed. For a single step, the $\max$ operation borrowed from the concept of max pooling is used to ensure that the agent can complete the exploration of the entire event as much as possible. In other words, the reward comes from two aspects, the first is the reward that the value of the single-step simulation is close to the observed value, and the other is the reward on the time step without collision in the single step, as shown in Figure. \ref{fig:max}.

\end{itemize}

\section{Model Benckmark Performance}
Evaluating the performance of a car-following model is essential for assessing its effectiveness in maintaining safe and efficient driving behavior. While there is a lack of universally standard evaluation metrics for this particular task, it is crucial to establish a set of criteria that effectively capture the key aspects of car-following. In this section, we present the metrics chosen as the standard evaluation criteria: mean square error of spacing and collision rate. A lower MSE score indicates a better fit of the model to the data and improved accuracy in predicting the spacing between vehicles. Additionally, high collision rates indicate that the model is not effective in avoiding collisions and requires further optimization. While a model with a low MSE score can accurately predict vehicle spacing, it may not perform well in avoiding collisions.
Therefore, using both MSE of spacing and collision rate helps to balance the trade-off between accuracy and safety when evaluating car-following models.

\subsection{MSE of spacing}
To assess the performance of five models on the above datasets, the MSE of spacing is utilized as the metric for assessment, as shown in Table. \ref{tab:results}. For one car-following event, the MSE of spacing can be expressed as:

\begin{equation}
    \operatorname{MSE}={\frac{1}{N} \sum_{i=1}^{N}\left({{S_{n - 1,n}}(t) - S_{n - 1,n}^{obs}(t)}\right)^{2}}
\end{equation}

where $N$ is the total number of observations, and $i$ is an observation index. Based on the benchmark conducted on five different car-following models, namely IDM, GHR, NN, LSTM, and DDPG, and evaluated on five distinct datasets, including NGSIM, HighD, Lyft, Waymo, and SPMD, as shown in Figure. \ref{fig:result}, it can be concluded that data-driven models, such as NN, LSTM, and DDPG, generally exhibit lower spacing errors compared to traditional models that rely on mathematical equations except for the HighD dataset. Data-driven models do not require calibration which is necessary for traditional models and can capture complex relationships between input variables, leading to more accurate predictions of car-following behavior. However, on the HighD dataset, which was collected on a highway, the average speed distribution of the car following events is highly concentrated in the high-speed range, as shown in Figure. \ref{fig:distribution}. GHR and IDM models achieve competitive results in MSE spacing with zero collision. Therefore, traditional models still have their advantages in a single scenario. The research suggests that while data-driven models generally outperform traditional models, it is important to carefully consider the scenario in question and the specific dataset being used when selecting the appropriate model. The findings highlight the advantages and limitations of different modeling approaches and provide valuable insights for the development of intelligent transportation systems.

\subsection{Collision rate}
The collision rate is a critical safety metric for car-following models, as even a single collision can result in severe consequences, such as injury or loss of life. According to the collision rate, as shown in Table. \ref{tab:results}, while the traditional models exhibit a higher spacing MSE, they are capable of achieving zero collision when calibrated using GA. DDPG model with a well-designed reward function also displays zero collision compared to NN and LSTM Models in the above datasets. 

Specifically, The largest time gap in the Lyft dataset, as shown in  Figure.  \ref{fig:distribution}, implies that the drivers in this dataset exhibit more conservative driving behavior, which can contribute to a safer driving environment. In addition, all models achieved zero collision rates in the Lyft dataset indicating that this dataset provides a valuable source of information for developing safe models.

\section{Discussion}
Based on the findings mentioned above, there are several points to discuss regarding to the model performance and potential future directions for car-following research.

\begin{itemize}
    \item \textbf{Safety}: Although data-driven models achieve lower MSE of spacing compared to traditional models, collisions can still occur. Therefore, it is desirable to develop car-following models that not only achieve lower spacing errors but also have zero collision rates, similar to the traditional models when properly calibrated. Incorporating the ability to avoid collisions into data-driven models would significantly enhance their safety and make them more suitable for real-world applications.

    \item\textbf{Interpretability and Generalization Ability}: Data-driven car-following models often lack interpretability and generalization ability. Each model requires specific tuning based on the dataset it is trained on. Models trained on a specific dataset, such as NGSIM, may not perform as well when directly applied to a different dataset, like Lyft. This discrepancy suggests that there are underlying variations and characteristics unique to each dataset that impact the car-following behavior. Therefore, it is necessary to develop car-following models that can generalize effectively across different datasets and driving scenarios. In future research, it would be valuable to investigate the performance of models trained on one dataset and tested on other datasets. This analysis would provide insights into the models' generalization capabilities and their ability to adapt to diverse driving conditions. Understanding how well models transfer their learned knowledge between datasets can help identify limitations and guide the development of more robust and adaptable car-following models. 

    

    \item \textbf{Expectations for Future Datasets}: To further enhance the performance and realism of car-following models, it is crucial to include additional factors in future datasets. For instance, incorporating data related to road conditions, and traffic signals would provide a more comprehensive understanding of the driving environment. Additionally, integrating information about surrounding vehicles and their behaviors would enable the models to account for complex interactions and make more accurate predictions. By incorporating these additional data sources, future datasets can better represent real-world driving scenarios, facilitating the development of more robust and effective car-following models.

    \item \textbf{More Advanced Algorithms}: The exploration of advanced machine learning techniques can significantly improve car-following models. For example, employing graph neural networks (GNNs) can help capture the intricate interactions between vehicles in a traffic network, leading to more accurate predictions of car-following behavior. Additionally, generative models can be used to synthesize new car-following events and expand the training data, enabling the models to learn from a larger and more diverse dataset. Moreover, leveraging meta-learning approaches can facilitate the development of adaptive car-following models that can quickly adapt to different datasets and driving conditions. These advanced algorithms have the potential to enhance the performance, adaptability, and scalability of car-following models, paving the way for more efficient and safe autonomous driving systems.
    
    \item \textbf{Mixed Traffic Flow}: A promising future direction for car-following research is to consider mixed traffic flow. As we transit towards autonomous driving systems, a mixed traffic flow scenario emerges where autonomous and human-driven vehicles share the road.  This involves developing car-following models that can effectively handle the interactions between autonomous and human-driven vehicles. Using datasets that include autonomous vehicles for training, particularly the Lyft and Waymo datasets provided in this benchmark, is crucial for the development of car-following models in mixed traffic flows. This approach can lead to more accurate modeling results, thereby improving the safety and efficiency of autonomous driving systems in these scenarios.
    

\end{itemize}

\section{Conclusion}

In this study, we propose FollowNet, the first benchmark of car-following behavior that includes five commonly used real-world datasets. Based on this benchmark,  we have analyzed the characteristics of car-following behavior with six dimensions. Additionally, we have evaluated the performance of different car-following models, using MSE of spacing and collision rate as evaluation metrics. The data-driven models, including NN, LSTM, and DDPG, outperform traditional models like IDM and GHR in terms of MSE of spacing. However, traditional models can achieve a zero collision rate when calibrated using GA, emphasizing the safety in car-following models is crucial. Especially, our proposed DDPGs\_Max model achieves
competitive performance on the benchmark with a smaller
MSE of spacing than traditional models such as IDM and
GHR in most datasets, as well as zero collision rate compared to NN
and LSTM models.

We believe that the establishment of a car-following benchmark with open access to data and source code will enable the development of more accurate and safe models, ultimately contributing to the advancement of microscopic traffic simulation models.

\bibliographystyle{IEEEtran}
\bibliography{ref}

\end{document}